%% file: main.tex
\documentclass[10pt, conference]{IEEEtran}
\IEEEoverridecommandlockouts

\usepackage{cite}
\usepackage[utf8]{inputenc}
\usepackage{enumitem}
\usepackage{url}
\usepackage{graphicx}
\usepackage{tikz}
\usepackage{caption}
\usepackage{subcaption}
\usepackage{multirow}
\usepackage{booktabs}
\usepackage{listings}
\usepackage{amsmath,amssymb,amsfonts}
\usepackage{comment}
\usepackage{indentfirst}
\usepackage{array} 
\usepackage{comment}
\usepackage[flushleft]{threeparttable}
\usepackage{makecell}
\usepackage{algorithmic}
\usepackage[norelsize, linesnumbered, ruled, vlined, boxed, commentsnumbered]{algorithm2e}
\usepackage[a4paper, total={184mm,239mm}]{geometry}
\usepackage[hidelinks, colorlinks=true, allcolors=blue]{hyperref} 
\selectfont
\usepackage{fancyhdr}
\usepackage{tabularx}
\usepackage{xcolor}
\SetKwComment{tcc}{\texttt{\scriptsize }}{}

\SetCommentSty{mycommfont}

\newcolumntype{L}[1]{>{\raggedright\let\newline\\\arraybackslash\hspace{0pt}}m{#1}}
\newcolumntype{C}[1]{>{\centering\let\newline\\\arraybackslash\hspace{0pt}}m{#1}}
\newcolumntype{R}[1]{>{\raggedleft\let\newline\\\arraybackslash\hspace{0pt}}m{#1}}

\begin{document}

\pagestyle{fancy} 
\fancyhf{} 
\renewcommand{\headrulewidth}{0pt} 
\renewcommand{\footrulewidth}{0pt} 
\cfoot{\thepage} 
\fancypagestyle{firstpage}{
  \fancyhf{}
  \renewcommand{\headrulewidth}{0pt}
  \renewcommand{\footrulewidth}{0pt}
  \fancyfoot[C]{\thepage}
  \fancyfoot[L]{\raisebox{-8pt}{\parbox[t]{0.90\textwidth}{\raggedright\scriptsize
  © 2026 IEEE. Personal use of this material is permitted. Permission from IEEE must be obtained
  for all other uses, in any current or future media, including reprinting/republishing this material
  for advertising or promotional purposes, creating new collective works, for resale or redistribution
  to servers or lists, or reuse of any copyrighted component of this work in other works.
  DOI: \texttt{}}}}
}


\title{
    \huge
    REvolution: An Evolutionary Framework for RTL Generation driven by Large Language Models\thanks{{\textsuperscript{*}These authors equally contributed to this work. This work has been accepted for publication at the 2026 Asia and South Pacific Design Automation Conference (ASP-DAC).
    The research was supported by nanomaterials development program through the National Research Foundation of Korea (NRF) (2022M3H4A1A04096496) funded by the Ministry of Science and ICT, Korea and LX Semicon.}}
}

\author{
\IEEEauthorblockN{Kyungjun Min\textsuperscript{*}, Kyumin Cho\textsuperscript{*}, Junhwan Jang, and Seokhyeong Kang} 
\IEEEauthorblockA{\textit{Department of Electrical Engineering, Pohang University of Science and Technology}\\
Pohang, Republic of Korea\\
\{kj.min, kmcho, jhjang17, shkang\}@postech.ac.kr}
}

\maketitle
\thispagestyle{firstpage}

\begin{abstract}
Large Language Models (LLMs) are used for Register-Transfer Level (RTL) code generation, but they face two main challenges: functional correctness and Power, Performance, and Area (PPA) optimization. 
Iterative, feedback-based methods partially address these, but they are limited to local search, hindering the discovery of a global optimum.
This paper introduces REvolution, a framework that combines Evolutionary Computation (EC) with LLMs for automatic RTL generation and optimization. 
REvolution evolves a population of candidates in parallel, each defined by a design strategy, RTL implementation, and evaluation feedback. 
The framework includes a dual-population algorithm that divides candidates into Fail and Success groups for bug fixing and PPA optimization, respectively.
An adaptive mechanism further improves search efficiency by dynamically adjusting the selection probability of each prompt strategy according to its success rate.
Experiments on the VerilogEval and RTLLM benchmarks show that REvolution increased the initial pass rate of various LLMs by up to 24.0 percentage points. 
The DeepSeek-V3 model achieved a final pass rate of 95.5\%, comparable to state-of-the-art results, without the need for separate training or domain-specific tools. 
Additionally, the generated RTL designs showed significant PPA improvements over reference designs. 
This work introduces a new RTL design approach by combining LLMs' generative capabilities with EC's broad search power, overcoming the local-search limitations of previous methods.
\end{abstract}


\input{content/1_intro/v_camera}

\input{content/2_background/v_camera}

\input{content/3_method/v_camera}

\input{content/4_expnrst/v_camera}

\input{content/5_conclusion/v_camera}

\newpage

\input{ref}
\end{document}

%% file: content/1_intro/v_camera.tex
\section{Introduction}

The growing complexity of modern integrated circuits has made manual Register-Transfer Level (RTL) design a significant bottleneck, prone to increased time, cost, and errors~\cite{PeiHM24}. 
In response, Large Language Models (LLMs) have recently been explored within Electronic Design Automation (EDA) research to generate Hardware Description Language (HDL) directly from natural-language specifications~\cite{ZhongDK23}.

However, directly applying LLMs to RTL design presents two key challenges.
First, functional correctness remains a major hurdle.
LLMs are predominantly trained on sequential software code, and thus struggle with the concurrent nature of HDL~\cite{QinXL25}.
This creates a performance ceiling: on the VerilogEval‑Human benchmark~\cite{LiuPK23}, state-of-the-art models such as \textit{Claude3.7‑Sonnet}, \textit{OpenAI o3}, and \textit{DeepSeek‑R1} attain only \(\sim\!80\%\) pass@1 accuracy~\cite{DengTL25}.
Second, standard LLMs lack awareness of post-synthesis metrics.
This results from their training being confined to HDL source code, excluding post-synthesis reports on Power, Performance, and Area (PPA)~\cite{ZhaoFL25}.
As a result, the generated designs are often PPA-suboptimal, necessitating numerous, inefficient design and synthesis iterations to meet specific targets.

To address these challenges, recent studies have introduced several promising approaches.
Frameworks such as MAGE~\cite{ZhaoZH24} and AIVRIL2~\cite{IslamSG25} employ iterative, agent-based workflows with EDA tool feedback to automatically correct functional errors. 
Similarly, to address PPA, approaches like PPA-RTL~\cite{ZhaoFL25} and VeriPPA~\cite{ThoratZL23} incorporate post-synthesis metrics as feedback. PPA-RTL, for instance, uses Reinforcement Learning (RL) to guide this optimization.
However, these feedback-driven, iterative methods share a common limitation: they perform a local search of the design space.
Their focus on refining an initial design candidate makes the final output's quality highly dependent on the starting point.
Consequently, if the initial implementation strategy is suboptimal, these methods risk getting trapped in a local optimum, failing to discover globally superior design alternatives. 
This necessitates a methodology capable of exploring a broad space of implementations, rather than merely refining a few design candidates.

To overcome these limitations, this paper proposes REvolution, a framework that integrates Evolutionary Computation (EC) with LLMs for automatic RTL code generation. 
At its core, REvolution evolves a population of design candidates, where each is defined by a design strategy (Thought), its corresponding RTL implementation (Code), and evaluation feedback. 
To guide this evolutionary process efficiently, the framework employs two key approaches.
First, a dual-population algorithm segregates candidates into `Fail' and `Success' populations, applying tailored strategies for bug fixing and PPA optimization, respectively. 
Second, an adaptive selection mechanism dynamically updates the probability of each prompt strategy based on its success, ensuring computational resources are prioritized effectively.
Using these approaches, REvolution can systematically explore the design space, moving beyond the local-search limitations of prior work to find functionally correct and PPA-optimized RTL code.
The contributions of this work are summarized as follows:

\begin{itemize}[noitemsep,topsep=0pt,leftmargin=*]

\item \textbf{Evolutionary Framework for RTL Generation and Optimization:} We introduce REvolution, the first framework to our knowledge that integrates the generative capabilities of LLMs with EC for RTL design. This approach moves beyond the local-search limitations of prior methods by enabling a parallel exploration of the design space for functionally correct and PPA-optimized RTL.

\item \textbf{Dual-Population Algorithm:} We propose a dual-population algorithm that segregates candidates into Success and Fail populations based on their functional correctness. This approach enhances search efficiency by applying heterogeneous evolutionary strategies tailored to the distinct goals of each population: bug fixing for the Fail population and PPA optimization for the Success population.
\vspace{0.1cm}
\item \textbf{Adaptive Prompt Strategy Selection:} We introduce a self-adaptive mechanism where the selection probability of each prompt strategy is dynamically updated based on its success. This allows REvolution to intelligently allocate computational resources by prioritizing more effective prompt strategies, significantly enhancing overall search efficiency.

\end{itemize}

\noindent
The remainder of this paper is organized as follows. Section~\ref{sec:background} covers the background. Section~\ref{sec:method} details the REvolution framework. Section~\ref{sec:exp} presents the experimental results. Finally, Section~\ref{sec:conclusion} concludes the paper.

%% file: content/2_background/v_camera.tex
\section{Background}
\label{sec:background}

\subsection{LLMs for RTL Generation}

Research on RTL code generation using LLMs can be broadly categorized into four key areas: data generation and augmentation~\cite{PeiHM24, ZhaoHL24, CuiYZ24, LiuFL24, ChangKN24, AkyashAK25, MinPP25, ThakurAP, LiuTZ25}, model and fine-tuning optimization~\cite{PeiHM24, QinXL25, DengTL25, ZhaoFL25, ZhaoHL24, LiuFL24, MinPP25}, prompt engineering~\cite{QinXL25, DengTL25, DeLorenzoTJ25}, and framework development~\cite{ZhaoZH24, IslamSG25, ThoratZL23, CuiYZ24, ChangRW23, ZuoLW25}. 
Each of these areas addresses different challenges in improving the quality and efficiency of RTL generation.

A major challenge in this field is the lack of high-quality, domain-specific datasets. 
To address this, OriGen~\cite{CuiYZ24} proposes a ``Code-to-Code'' augmentation technique that refines open-source code using a teacher LLM based on newly generated descriptions. 
VeriLogos~\cite{MinPP25} modifies the Abstract Syntax Tree (AST) of the code to create new synthetic data.
In addition to dataset enhancement, optimizing the model and fine-tuning it to embed a deeper understanding of hardware-specific characteristics is crucial. 
PPA-RTL~\cite{ZhaoFL25} uses RL to directly optimize the model for post-synthesis PPA metrics. 
BetterV~\cite{PeiHM24} combines a generative model with a discriminator that guides the output to optimize for specific EDA tasks.

Prompt engineering is crucial for enhancing an LLM's reasoning without the need for expensive retraining. 
Abstractions-of-Thought~\cite{DeLorenzoTJ25} introduces a structured three-stage prompting framework (`classification - intermediate representation - pseudocode') to guide the LLM’s reasoning hierarchically. 
ReasoningV~\cite{QinXL25} uses an adaptive reasoning mechanism that adjusts the inference depth based on problem complexity to improve efficiency.
Finally, framework development enhances RTL generation through systematic methods such as multi-agent or iterative feedback loops. 
MAGE~\cite{ZhaoZH24} employs a multi-agent system where specialized agents handle RTL generation, testbench creation, and debugging. 
VeriPPA~\cite{ThoratZL23} uses a multi-round feedback loop that incorporates outputs from simulators and PPA reports into the next prompt, iteratively refining the code until it meets the target.

These approaches have improved the accuracy of LLM-based RTL generation, with some even advancing to PPA optimization. However, they lack a methodology for exploring a broad design space. Fine-tuned models may be biased toward specific answers, and feedback-based frameworks focus on refining a few solutions rather than exploring many alternatives. To address this gap, REvolution combines LLMs and EC to efficiently explore the design space for functionally correct and PPA-optimal RTL designs.

\subsection{Integrating LLMs and Evolutionary Computation}

Recently, combining LLM code generation with EC search methods has become a key research area~\cite{Romera-ParedesBN24, NovikovVE25, LiuTY24}. This approach focuses on discovering verifiable algorithms that are challenging for standalone LLMs to generate, by searching for programs (or functions) that produce solutions, rather than the solutions themselves.
FunSearch~\cite{Romera-ParedesBN24} evolves a specific function within a program skeleton. AlphaEvolve~\cite{NovikovVE25} extends this by evolving entire code files, enabling the modification of complex algorithms. 
In contrast, EoH~\cite{LiuTY24} evolves not only code but also `Thoughts' (natural language descriptions of heuristic ideas). 
This dual representation, combined with systematic prompt strategies, aims to generate high-performance heuristics more efficiently.

However, the direct application of these approaches to the RTL generation domain faces distinct challenges. 
Whereas a typical heuristic search is evaluated on a single-objective score, RTL generation is a multi-objective optimization problem that requires satisfying functional correctness while considering the complex trade-offs between PPA.
Therefore, to address this research gap, REvolution enhances search efficiency by employing a dual-population strategy that separates individuals based on their functional correctness. 
Furthermore, it directly utilizes evaluation feedback in the evolutionary process to effectively guide the exploration of the design space.

\input{table/term}

%% file: table/term.tex
\begin{table}[t]
\centering
\caption{Summary of Terminology and Hyperparameter}
\label{tab:terminology}
\begin{tabularx}{\columnwidth}{lX}
\toprule
\textbf{Term/Symbol} & \textbf{Description} \\
\midrule
\multicolumn{2}{c}{\textbf{Core Concepts}} \\
\midrule
Population & Set of individuals, representing design candidates. \\
Individual & Single design candidate, represented as a (Thought, Code, Feedback) tuple. \\
Fitness & Score measuring an individual's quality. \\
Offspring & New individual created from selected parents. \\
Parent(s) & Selected individuals from the current population that produce offspring. \\
Thought & High-level design strategy in natural language. \\
Code & Verilog RTL implementation of a Thought. \\
Feedback & LLM-generated analysis of Code's evaluation results. \\
\midrule
\multicolumn{2}{c}{\textbf{Evolutionary Algorithm}} \\
\midrule
$N$ & Total number of individuals in the population. \\
$\lambda$ & Number of offspring generated per generation. \\
$n$ & Number of elite individuals preserved per PPA metric. \\
$G$ & Maximum number of generations for termination. \\
$\alpha, \beta, \gamma$ & Weights for the PPA objectives in the fitness function. \\
\midrule
\multicolumn{2}{c}{\textbf{Adaptive Strategy Selection}} \\
\midrule
$R$   & Reward for a given prompt strategy. \\
$Q(i)$ & Average reward (action-value) for strategy $i$. \\
$c$ & Exploration parameter for the UCB algorithm. \\
$\tau$  & Softmax selection temperature. \\
$T$ & Total number of prompt strategy selections. \\
$k_i$  & Selection count for a specific strategy $i$. \\
$M$   & Number of available prompt strategies. \\
\bottomrule
\end{tabularx}
\vspace{-0.2cm}
\end{table}

%% file: content/3_method/v_camera.tex
\input{figure/flow}

\section{Methodology}
\label{sec:method}

\subsection{Overview}
REvolution integrates LLMs with EC to automatically generate functionally correct and PPA-optimized RTL code. 
Exploring and evolving a population of design candidates in parallel reduces dependency on the initial design, mitigating the risk of local optima and enabling efficient exploration of a broader design space.
Figure~\ref{fig:overall_flow} illustrates the overall flow of the framework. 
The process begins with an Initialization, where an LLM creates an initial population of diverse design candidates. 
This is followed by the Evolutionary Loop, which evolves the design candidates by evaluating them, generating offspring through prompt strategies, and selecting the fittest for the next generation.
This cycle repeats until a termination condition is met, and the framework returns the best RTL code found.
The terminology and hyperparameters of our framework are summarized in Table~\ref{tab:terminology}.
The following sections detail the core components of our framework, including the evolutionary representation~\ref{sec:3.2}, the dual-population algorithm~\ref{sec:3.3}, prompt strategies~\ref{sec:3.4}, and adaptive prompt strategy selection~\ref{sec:3.5}, before concluding with a description of the complete REvolution algorithm~\ref{sec:3.6}.

\input{figure/individual}

\subsection{Evolutionary Representation}
\label{sec:3.2}

The fundamental unit of evolution is the individual, which is defined as a (Thought, Code, Feedback) tuple (Figure~\ref{fig:individual}). 
The Thought is a natural language description representing the high-level design strategy. 
The Code is the Verilog RTL implementation derived from the Thought. 
The Feedback is a natural language summary dynamically generated by an LLM that analyzes the Code's evaluation results.
If the Code fails simulation, the LLM analyzes error logs to generate bug-fixing feedback. 
If the Code is correct, the LLM reviews PPA reports and suggests possible improvements.

Furthermore, to quantitatively evaluate each individual, we define a fitness score. 
Functionally incorrect individuals are assigned a uniform score of $-\infty$, ensuring they are always ranked below any successful candidate during survivor selection.
Conversely, for functionally correct individuals, the fitness score is calculated as follows:

\begin{equation*}
\mbox{\footnotesize $ \displaystyle
\text{F}_{gen} = \alpha \left( \frac{P_{ref} - P_{gen}}{P_{ref}} \right) + \beta \left( \frac{A_{ref} - A_{gen}}{A_{ref}} \right) + \gamma \left( \frac{T_{ref} - T_{gen}}{T_{ref}} \right)
$}
\end{equation*}

\noindent
The variables $P$, $A$, and $T$ represent power, area, and effective clock period for the generated ($gen$) and reference ($ref$) designs, with each objective weighted by the user-defined coefficients $\alpha$, $\beta$, and $\gamma$.

\subsection{Dual-Population Algorithm}
\label{sec:3.3}

In each generation, the dual-population algorithm divides the population into two sub-populations: the Fail population (functionally incorrect individuals) and the Success population (functionally correct individuals).
By segregating the population, we apply heterogeneous strategies (detailed in Section \ref{sec:3.4} and \ref{sec:3.5}) to target the distinct goals of each sub-population: functional correctness for the Fail population and PPA optimization for the Success population.
The dual-population algorithm adjusts offspring generation based on sub-population size, directing more effort to the larger sub-population to address the pressing challenge.
The number of offspring is directly proportional to the relative size of each sub-population in the current generation. 
For example, with $\lambda$ total offspring and a 7:3 Fail-to-Success ratio, the Fail and Success populations will generate 0.7$\lambda$ and 0.3$\lambda$ offspring, respectively.

\subsection{Prompt Strategies}
\label{sec:3.4}

The generation of new individuals is driven by a series of prompts that serve as the genetic operators for our framework.
These prompts are categorized into two types. 
The Initial Prompt, containing only the user-provided functional description, is used just once during the Initialization stage to generate the initial population. 
In contrast, the Evolutionary Prompts are used repeatedly within the Evolutionary Loop to create new offspring from existing individuals.
Each Evolutionary Prompt combines the objective, functional description, and the (Thought, Code, Feedback) from selected parents to instruct the LLM to generate a new Thought and Code pair for a new individual (Figure~\ref{fig:prompt}). 
The following are descriptions of six prompt strategies:

\begin{itemize}[noitemsep,topsep=0pt,leftmargin=*]

\item \textbf{Fix:} Corrects functional errors in a single parent's Thought and Code based on the provided Feedback.

\item \textbf{Simplify:} Reduces the complexity of a single parent's Thought and Code to escape from complex or erroneous design states.

\item \textbf{Explore:} Generates a completely new and different Thought and its corresponding Code from a single parent to enhance the diversity of the population.

\item \textbf{Refactor:} Maintains the original Thought from a single parent but re-implements the Code in a structurally different way.

\item \textbf{Improve:} Makes general enhancements to a single parent's Thought and Code to improve its overall quality and performance.

\item \textbf{Fusion:} Merges the Thoughts and Codes of two successful parent individuals, aiming to create a new offspring that inherits the strengths of both.

\end{itemize}

\noindent
These prompt strategies are applied heterogeneously to the Fail and Success populations. 
Both groups utilize Simplify, Explore, Refactor, and Improve. 
However, the Fail population additionally uses the Fix operator to correct errors, while the Success population instead uses the Fusion operator to combine proven, successful designs.

\subsection{Adaptive Prompt Strategy Selection}
\label{sec:3.5}
The prompt strategy selection is based on a multi-armed bandit problem, where each strategy represents an ``arm''~\cite{SuttonB98}. 
Reward ($R$) is given for generating a functionally correct design (Fail population) or improving fitness score (Success population). 
The Upper Confidence Bound (UCB) algorithm calculates the strategy score for each strategy $i$ as follows:

\begin{equation*}
\label{eq:ucb_short}
\text{Score}(i) = Q(i) + c \sqrt{\frac{\ln T}{k_i}}
\end{equation*}

\noindent
Here, $Q(i)$ is the average reward for strategy $i$, $k_i$ is its selection count, $T$ is the total number of selections, and $c$ is the exploration parameter. 
After the $k_i$-th use of a strategy, its action-value $Q(i)$ is updated with an incremental average:

\begin{equation*}
\label{eq:q_update_short}
Q_{k_i+1}(i) = Q_{k_i}(i) + \frac{1}{k_i} [R_{k_i} - Q_{k_i}(i)]
\end{equation*}

\noindent
where $R_{k_i}$ is the reward received for the $k$-th use of the strategy $i$.
To prevent premature convergence, we use the scores to generate a probability distribution via the softmax function:

\begin{equation*}
\label{eq:softmax_short}
P(i) = \frac{\exp(\text{Score}(i) / \tau)}{\sum_{j=1}^{M} \exp(\text{Score}(j) / \tau)}
\end{equation*}

\noindent
where $\tau$ is the temperature parameter and $M$ is the number of available prompt strategies. 
This hybrid UCB-Softmax approach facilitates both adaptive learning and effective exploration.





\subsection{The REvolution Algorithm}
\label{sec:3.6}

The REvolution framework consists of three main stages: Initialization, the Evolutionary Loop, and Termination (Figure~\ref{fig:overall_flow}).

In the Initialization stage, an initial population of size $N$ is generated using the Initial Prompt. 
Each individual is then evaluated through simulation and synthesis. 
Based on the simulation result, individuals are classified into either a Success population or a Fail population. 
In this process, the fitness score and feedback for each individual are also computed and stored.

Following initialization, the framework evolves the population by iterating through the Evolutionary Loop. 
This loop comprises three steps: offspring generation, offspring evaluation, and survivor selection.
The first step, offspring generation, produces $\lambda$ new individuals. 
A prompt strategy is initially selected via an adaptive mechanism. 
Subsequently, parents are chosen from their respective sub-populations. 
For the Fail population, where individuals have identical fitness scores, parents are selected randomly. 
For the Success population, however, roulette wheel selection is employed, making an individual's selection probability proportional to its fitness.
In the second step, offspring evaluation, each new offspring is assessed to compute its fitness score and feedback.
This evaluation measures the success of the applied prompt strategy, updating its selection probability for future generations.
The final step, survivor selection, determines the $N$ individuals for the succeeding generation. 
Initially, the top $n$ individuals for each of the three PPA metrics from the parent population are preserved.
The remaining $N - 3n$ slots are then populated by the fittest individuals from a combined population of parents and offspring.
The new population is re-segregated into Fail and Success sub-populations based on functional correctness for the next iteration.

The Termination stage is reached after a predefined number of generations ($G$). 
Finally, the framework outputs the Thought, Code, and PPA results of the individual with the highest fitness score recorded across all generations.

\input{figure/prompt}

%% file: figure/flow.tex
\begin{figure*}[th]
    \centering
    \includegraphics[width=0.9\linewidth]{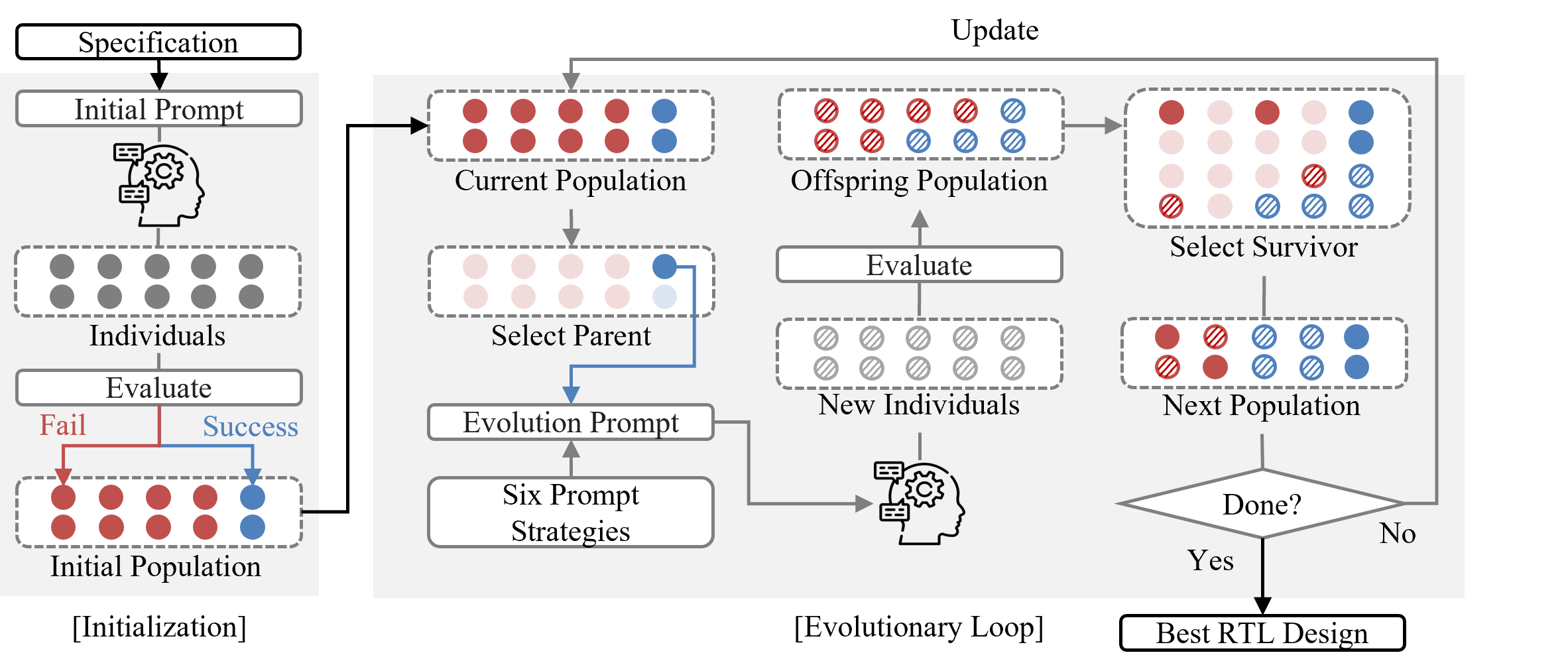}
    \caption{
    Overall framework of the REvolution. The framework initializes a population from a specification (functional description) and refines it through an evolutionary loop of offspring generation, evaluation, and survivor selection until termination.}
    \label{fig:overall_flow}
    \vspace{-0.5cm}
\end{figure*}

%% file: figure/individual.tex
\begin{figure}[t]
    \centering
    \includegraphics[width=0.9\linewidth]{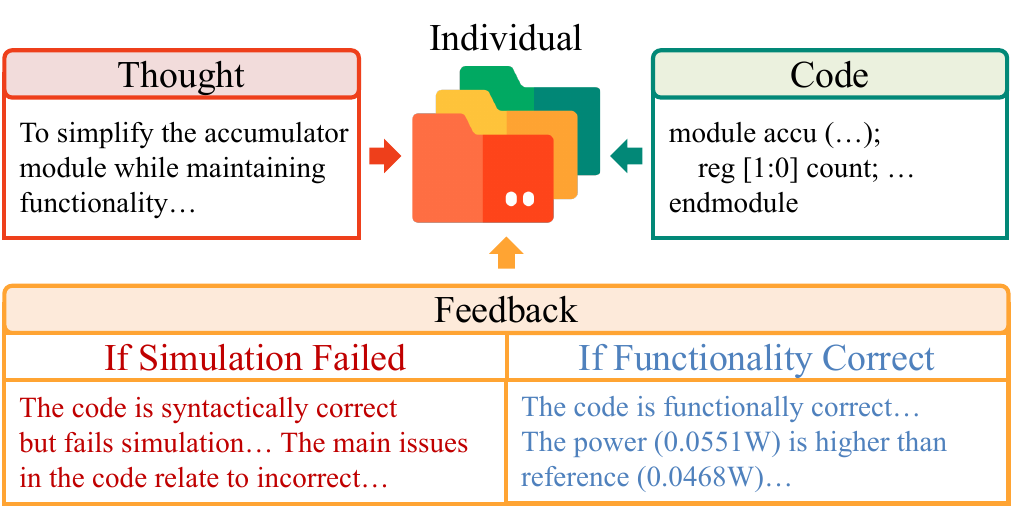}
    \caption{Structure of Individual. Individual is a (Thought, Code, Feedback) tuple, with Feedback generated based on the Code's functional correctness.}
    \label{fig:individual}
    \vspace{-0.3cm}
\end{figure} 

%% file: figure/prompt.tex
\begin{figure}[t]
    \centering
    \includegraphics[width=0.9\linewidth]{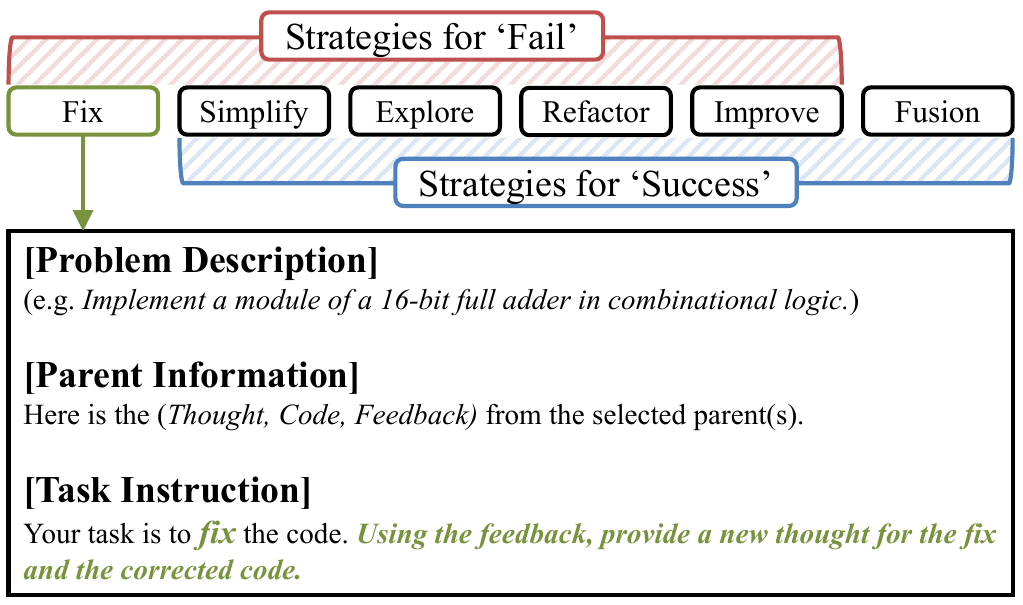}
    \caption{Example of an Evolutionary Prompt. The available prompt strategies vary depending on the population, and each prompt is constructed from a general template.}
    \label{fig:prompt}
    \vspace{-0.4cm}
\end{figure}

%% file: content/4_expnrst/v_camera.tex
\section{Experimental Results}
\label{sec:exp}

\input{table/result}

\subsection{Experimental Setup}

We evaluated our framework using GPT-4.1-mini (2025-04-14)~\cite{GPT-4.1-mini}, DeepSeek-V3 (0324)~\cite{Deepseek}, and Llama-3.3-70B~\cite{Llama}, all accessed via API with a temperature of 1.0 and top-p of 0.95.
The hyperparameters were configured as follows: a population size ($N$) of 10, an offspring count ($\lambda$) of 10 per generation, a maximum of 20 generations ($G$)\footnote{These hyperparameter values were set to balance search effectiveness with our computational budget, which is primarily constrained by LLM API costs.}, an elite count ($n$) of 1, a reward ($R$) of 1, and exploration hyperparameters ($c$, $\tau$) set to (2.0, 1.0). 
Fitness weights ($\alpha$, $\beta$, $\gamma$) were set to (1/2, 1/2, 0) for combinational circuits and (1/3, 1/3, 1/3) for sequential ones.
The experiments were conducted on the VerilogEvalV2 (Spec-to-RTL)~\cite{PinckneyBL24} and RTLLM-2.0 benchmarks~\cite{LiuLF24}.
We used Icarus Verilog (iverilog)~\cite{iverilog} for simulation and Yosys~\cite{yosys} with the Nangate 45nm PDK~\cite{Nangate45} for synthesis, applying a uniform 0.01 ns clock period to all designs.
Performance was measured by Pass Rate and PPA Improvement.
The Pass Rate indicates the functional correctness of the final design.
The PPA Improvement measures the PPA gain over the reference design, reported only for designs with over 50 gates\footnote{The VerilogEvalV2 benchmark has an average gate count of approx. 113 (max 4,123), while RTLLM-2.0 has an average of approx. 288 (max 4,325). Our PPA analysis considers 26 designs from VerilogEvalV2 and 24 from RTLLM-2.0 that meet the $>$50 gate count threshold.}\footnote{Four designs from RTLLM-2.0 (adder\_pipe\_64bit, multi\_booth\_8bit, float\_multi, and synchronizer) were excluded from the PPA analysis as their reference designs failed to synthesize with our toolchain.}\footnote{We observed that some designs pass pre-synthesis simulation but fail functional checks post-synthesis. Therefore, while the Pass Rate is based on pre-synthesis results, PPA improvement is only reported for designs confirmed to be functionally correct after synthesis.}.
To ensure reproducibility, all scripts, logs, and results are publicly available at our repository~\cite{REvolution}.

\subsection{Performance Analysis and Comparison}

In this section, we quantitatively analyze the performance of the REvolution framework with various LLMs and compare it against state-of-the-art baseline method. 
As shown in Table~\ref{tab:result}, our evolutionary approach demonstrates significant effectiveness. 
REvolution with DeepSeek-V3 achieves the highest final Pass Rate of 95.5\% on the VerilogEval benchmark. 
Notably, the evolutionary loop consistently improves the initial Pass Rate across all models and benchmarks. 
For instance, with Llama-3.3-70B on the RTLLM benchmark, the Pass Rate increased from 60.0\% to 84.0\%, a substantial improvement of 24.0 percentage points. 
Functionally correct designs also showed considerable average PPA improvements.
The high improvement rates, such as the 67.0\% power reduction with Llama-3.3-70B on RTLLM, are attributable to the non-PPA-optimized nature of the reference designs; for transparency, all detailed synthesis reports and logs are available in our public repository. 
The average runtime generally correlates with the size and API response speed of the underlying LLM.

We chose VerilogCoder~\cite{HoRK25} as the primary SOTA baseline for comparison because it reports a final Pass Rate from a single framework execution, which aligns with our methodology, unlike methods that use pass@k metrics. 
VerilogCoder (GPT-4-Turbo) achieves a notable 94.2\% Pass Rate on VerilogEval, a performance that is comparable to our best result of 95.5\% with REvolution (DeepSeek-V3). 
On the Llama architecture, however, a significant performance gap is observed: while VerilogCoder (Llama-3-70B) reaches a 67.3\% Pass Rate, our REvolution (Llama-3.3-70B) elevates this to 88.5\% on the same benchmark, demonstrating how our approach enhances the base model's capability.
This performance difference stems from the distinct methodologies employed. 
VerilogCoder uses a task and circuit relation graph for planning and a custom AST-based waveform tracing tool for debugging. 
In contrast, REvolution achieves its results without such specialized tools, instead using a more general and flexible evolutionary approach guided solely by prompt strategies. 
This highlights a key advantage of our framework: it reduces the need for intricate, domain-specific tool development while still achieving state-of-the-art results.

\input{figure/case}

\subsection{Case Study: 8-bit Signed Adder}

This section presents a case study to provide a detailed, step-by-step illustration of how the REvolution framework operates. 
We analyze the results for the VerilogEval Prob033 problem, which specifies an 8-bit 2's complement adder with overflow detection, using DeepSeek-V3.
The evolutionary process and its results are summarized in Figure~\ref{fig:case} (Evolutionary Trajectory) and Figure~\ref{fig:ppa} (PPA Distribution\footnote{Since the design is a combinational circuit, this scatter plot visualizes the power-area distribution.}). 

The process started with a functionally correct design, but with a negative fitness score of -0.046, indicating inferior PPA compared to the reference design.
The first successful improvement strategy targeted the overflow detection logic.
The LLM's Thought process identified a more efficient XOR-based method, improving the fitness score to 0.0.
The next improvement step guided the LLM to implement a structural carry-lookahead architecture.
This demonstrates that REvolution can shift a design from behavioral to structural modeling, leading to a significant fitness increase to 0.293.
Finally, a Fusion strategy was applied to refine the design further, combining hierarchical carry computation ideas to create a balanced 2-bit block structure, achieving a peak fitness of 0.384. 
The final evolved design uses a structural carry-lookahead adder, contrasting with the reference design's high-level behavioral implementation.

Beyond the trajectory of a single design, the PPA scatter plot illustrates the broader design space explored by the evolutionary loop. 
The plot, with designs categorized into three generation groups (Generations 0-4, 5-11, and 12-20), shows a clear trend of PPA optimization. 
As generations progress, the centroid of the population moves towards the lower-left corner, indicating simultaneous improvements in both power and area. 
Furthermore, the distribution of all individuals versus the `Top 5' shows that while the framework maintains diversity, the selection process converges towards superior solutions.
In summary, this case study shows that REvolution can optimize logic and implement fundamental design changes, guiding the population towards PPA-optimized regions of the design space.

\subsection{Ablation Study: Impact of the Evolutionary Loop}

To assess whether the framework's performance gains are attributable to the evolutionary process or simply to an increased volume of design generation, we conducted an ablation study. 
We compared REvolution against a baseline (``Baseline (Llama-3.3-70B)'' in Table~\ref{tab:result}) that generates 200 designs using an Initial Prompt.
This number matches the total quantity of designs in a standard REvolution run (10 offspring $\times$ 20 generations), ensuring an identical computational budget for LLM calls between the two experiments.

The results show a distinct advantage for the evolutionary method in both functional correctness and PPA optimization. 
As detailed in Table~\ref{tab:result}, REvolution (Llama-3.3-70B) achieved higher final Pass Rates on both RTLLM (84.0\% vs. 70.0\%) and VerilogEval (88.5\% vs. 79.5\%) compared to the baseline. 
The performance gap is also evident in the PPA metrics. 
For instance, REvolution yielded a 67.0\% power reduction on RTLLM, substantially higher than the baseline's 37.3\%. 
Similarly, its power and clock improvements on VerilogEval (47.3\% and 48.2\%) significantly exceeded the baseline results (7.7\% and 2.3\%).
In conclusion, this study confirms that REvolution's performance stems from its structured, iterative evolution process, not merely from high-volume sampling. 
The framework's effectiveness is twofold: the higher Pass Rate demonstrates that the guided prompt strategies systematically correct functional errors, achieving valid solutions beyond the reach of random chance.
Concurrently, the superior PPA metrics indicate that the fitness-driven selection process actively steers the design population towards optimal regions of the PPA space. 
The evolutionary loop is, therefore, the key mechanism enabling these systematic enhancements, yielding results not achievable through an equivalent volume of independent sampling.

\input{figure/ppa}

%% file: table/result.tex
\begin{table*}[t]
\centering
\caption{Performance Comparison of REvolution and Baselines}
\label{tab:result}
\begin{tabular}{l l | c c c | c c c | c} 
\toprule
\multirow{2}{*}{\textbf{Model}} & \multirow{2}{*}{\textbf{Benchmark}} & \multicolumn{3}{c|}{\textbf{Pass Rate (\%)}} & \multicolumn{3}{c|}{\textbf{Avg PPA Improv. (\%)}} & \multirow{2}{*}{\thead{Avg Runtime \\ (s)}} \\
\cmidrule(r){3-5} \cmidrule(r){6-8}
& & \textbf{Init*} & \textbf{Final} & \textbf{($\Delta$)} & \textbf{Area} & \textbf{Power} & \textbf{Eff. Clk} & \\
\midrule
\multirow{2}{*}{\makecell[l]{REvolution \\ (DeepSeek-V3)}} & RTLLM & 64.0 & 88.0 & \textcolor{blue}{\textbf{+24.0}} & 27.9 & 51.4 & 23.6 & 1990 \\
& VerilogEval & 83.3 & 95.5 & \textcolor{blue}{\textbf{+12.2}} & 5.4 & 24.5 & 13.6 & 1515 \\
\midrule
\multirow{2}{*}{\makecell[l]{REvolution \\ (GPT-4.1-mini)}} & RTLLM & 70.0 & 86.0 & \textcolor{blue}{\textbf{+16.0}} & 26.5 & 44.2 & 15.3 & 1165 \\
& VerilogEval & 82.1 & 94.2 & \textcolor{blue}{\textbf{+12.1}} & 6.0 & 45.6 & 17.1 & 785 \\
\midrule
\multirow{2}{*}{\makecell[l]{REvolution \\ (Llama-3.3-70B)}} & RTLLM & 60.0 & 84.0 & \textcolor{blue}{\textbf{+24.0}} & 28.7 & 67.0 & 13.2 & 1605 \\
& VerilogEval & 67.3 & 88.5 & \textcolor{blue}{\textbf{+21.2}} & 4.3 & 47.3 & 48.2 & 1459 \\
\cmidrule(r){1-9}
\multirow{2}{*}{\makecell[l]{Baseline \\ (Llama-3.3-70B)}} & RTLLM & 70.0 & - & - & 15.9 & 37.3 & 2.4 & 386 \\
& VerilogEval & 79.5 & - & - & 3.4 & 7.7 & 2.3 & 658 \\
\midrule
\makecell[l]{VerilogCoder~\cite{HoRK25} \\ (Llama-3-70B)} & VerilogEval & N/A & 67.3 & - & N/A & N/A & N/A & N/A \\
\midrule
\makecell[l]{VerilogCoder~\cite{HoRK25} \\ (GPT-4-Turbo)} & VerilogEval & N/A & 94.2 & - & N/A & N/A & N/A & N/A \\
\bottomrule
\end{tabular}
\par
\vspace{0.1cm}
\small *Init refers to the pass rate of the initial population generated before the evolutionary loop begins.
\vspace{-0.2cm}
\end{table*}

%% file: figure/case.tex
\begin{figure*}[th]
    \centering
    \includegraphics[width=0.8\linewidth]{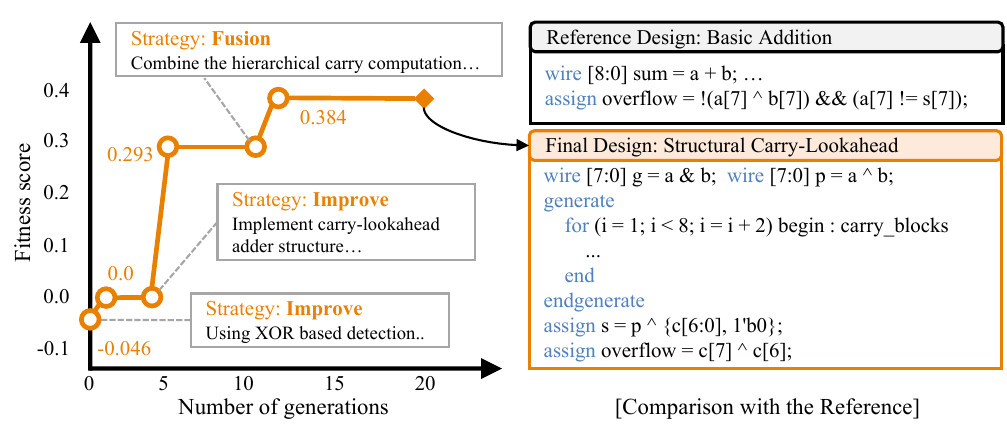}
    \caption{
     Evolutionary Trajectory of DeepSeek-V3 on VerilogEval Prob033 problem.}
    \label{fig:case}
\vspace{-0.3cm}
\end{figure*}

%% file: figure/ppa.tex
\begin{figure}[t]
    \centering
    \includegraphics[width=0.9\linewidth]{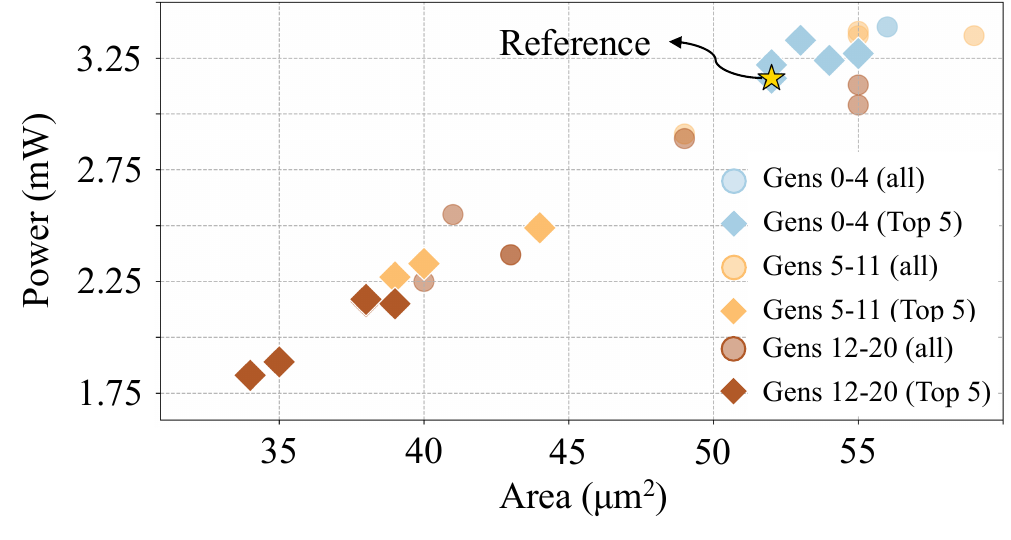}
    \caption{Power-area distribution of design candidates in the evolutionary process of DeepSeek-V3 on VerilogEval Prob033.}
    \label{fig:ppa}
    \vspace{-0.4cm}
\end{figure} 

%% file: content/5_conclusion/v_camera.tex
\section{Conclusion}
\label{sec:conclusion}

\noindent
This paper addresses the key challenges in LLM-based RTL generation: functional correctness and PPA optimization. 
To overcome these, we introduced REvolution, a novel framework that combines the generative power of LLMs with the robust search capabilities of EC.
Our approach evolves a population of design candidates, each comprising a Thought, Code, and Feedback, using a dual-population algorithm to apply specialized strategies for bug fixing and PPA improvement.
The efficiency of this evolutionary process is further enhanced by an adaptive prompt strategy selection mechanism.

Our experimental results demonstrate the effectiveness of the REvolution framework. 
On the VerilogEval and RTLLM benchmarks, REvolution significantly improved the initial pass rates of various LLMs by up to 24.0 percentage points, achieving a final pass rate of 95.5\% that is competitive with state-of-the-art methods.
Critically, these results were achieved without domain-specific model fine-tuning or complex external tools, highlighting the architectural advantage of our approach. 
Moreover, the evolved designs showed significant PPA improvements over the reference designs, confirming REvolution's ability to optimize both functional correctness and hardware efficiency.
In conclusion, REvolution offers a new approach for automated RTL design, enabling the discovery of highly optimized hardware solutions through the parallel evolution of multiple candidates.